\documentclass[a4paper,conference]{IEEEtran}
\ifCLASSINFOpdf
\else
\fi
\usepackage{graphics} 
\usepackage{epsfig} 
\usepackage{times} 
\usepackage{cite}
\usepackage{amsmath,amssymb,amsfonts}
\usepackage{algorithmic}
\usepackage[graphicx]{realboxes}
\usepackage{textcomp}
\usepackage[table,xcdraw]{xcolor}
\usepackage{multirow}
\usepackage{array}
\usepackage{tikz}
\usepackage{siunitx}
\usepackage{lscape}
\usepackage{adjustbox}
\usepackage{tabularx}
\usepackage{makecell}
\usepackage{mleftright}
\usepackage{etoolbox}
\usepackage{xcolor}
\usepackage[normalem]{ulem}
\usepackage{tablefootnote}
\usepackage{verbatim}
\usepackage{graphicx}
\usepackage{gensymb}

\DeclareMathOperator*{\argmin}{argmin}

\IEEEoverridecommandlockouts

\begin{document}
%
\title{Neural Architecture Adaptation for Object Detection by Searching Channel Dimensions \\ and Mapping Pre-trained Parameters \thanks{This work was supported by NCSOFT, in part by Institute of Information \& communications Technology Planning \& Evaluation (IITP) grant funded by the Korea government (MSIT) (No. 2019-0-00079,  Artificial Intelligence Graduate School Program (Korea University)).}}

\author{\IEEEauthorblockN{Harim Jung}
\IEEEauthorblockA{Dept. Artificial Intelligence \\
Korea University \\
Seoul, Republic of Korea\\
Email: hr\_jung@korea.ac.kr}
\and
\IEEEauthorblockN{Myeong-Seok Oh}
\IEEEauthorblockA{Dept. Computer Engineering \\
Korea University\\
Seoul, Republic of Korea\\
Email: ms\_oh@korea.ac.kr}
\and
\IEEEauthorblockN{Cheoljong Yang}
\IEEEauthorblockA{Vision AI Lab, AI Center\\
NCSOFT\\
Seongnam, Republic of Korea\\
Email: cjyang@ncsoft.com}
\and
\IEEEauthorblockN{Seong-Whan Lee}
\IEEEauthorblockA{Dept. Artificial Intelligence \\
Korea University \\
Seoul, Republic of Korea\\
Email: sw.lee@korea.ac.kr}}




\maketitle
\begin{abstract}
Most object detection frameworks use backbone architectures originally designed for image classification, conventionally with pre-trained parameters on ImageNet. However, image classification and object detection are essentially different tasks and there is no guarantee that the optimal backbone for classification is also optimal for object detection. Recent neural architecture search (NAS) research has demonstrated that automatically designing a backbone specifically for object detection helps improve the overall accuracy. In this paper, we introduce a neural architecture adaptation method that can optimize the given backbone for detection purposes, while still allowing the use of pre-trained parameters. We propose to adapt both the micro- and macro-architecture by searching for specific operations and the number of layers, in addition to the output channel dimensions of each block. It is important to find the optimal channel depth, as it greatly affects the feature representation capability and computation cost. We conduct experiments with our searched backbone for object detection and demonstrate that our backbone outperforms both manually designed and searched state-of-the-art backbones on the COCO dataset.
\end{abstract}


%
\IEEEpeerreviewmaketitle

\section{INTRODUCTION}
In recent years, there has been a growing interest in neural architecture search (NAS), which automates the process of designing neural network architectures. 
NAS is useful since the manual design of neural architectures heavily relies on human experts' limited experience and prior knowledge, which may lead to tedious trial and error. 
The automatically searched architectures have shown highly competitive performance compared to manually designed architectures in various tasks\cite{760572, YANG20073120, lee1990translation, lee2003pattern, lim2000text, xi2002facial}, including image classification \cite{zoph2018learning, fang2020densely, zoph2016neural, jordao2021stage}, language modeling \cite{pham2018efficient, liu2018darts}, object detection \cite{fang2020fna, chen2019detnas, zhang2021eod, yao2021joint}, and semantic segmentation \cite{liu2019auto, chen2018searching}. 
Object detection is a task of localizing and classifying various scales of objects in a given image and NAS for object detection can be mainly divided into three categories: backbone \cite{chen2019detnas, fang2020fna}, neck \cite{ghiasi2019fpn, xu2019auto}, and head search \cite{chen2020mnasfpn, wang2020fcos}. The search can be conducted through various strategies and most previous works have used reinforcement learning \cite{chen2020mnasfpn, xu2019auto, ghiasi2019fpn}, evolutionary algorithms \cite{chen2019detnas, guo2020powering}, or gradient-based learning \cite{peng2019nats, fang2020densely, fang2020fna}. 

Most existing object detectors directly utilize the backbone designed for image classification and its pre-trained parameters on ImageNet \cite{deng2009imagenet}. However, image classification and object detection are essentially different tasks and there is no guarantee that the optimal backbone for image classification is also optimal for object detection. Therefore, it is necessary to specifically search for a backbone architecture that can extract features appropriate for detecting and classifying multiple objects of different scales. One of the barriers of applying NAS to object detection is the pre-training cost, as backbone pre-training is still a necessary but costly procedure in object detection, in order to achieve faster convergence and higher accuracy. The search process can incur extra costs from pre-training the supernet as in DetNAS \cite{chen2019detnas}, which encompasses all candidate architectures. Therefore, instead of searching for an entire network architecture from scratch as in \cite{chen2019detnas, liu2019auto, fang2020densely}, we adopt an efficient function-preserving neural architecture adaptation method to search the optimal backbone for object detection. 

For the first time, we propose to adapt an existing backbone network in both the macro- and micro-level, while still permitting the use of ImageNet pre-trained parameters. We adjust the micro-architecture by modifying the specific operations and the macro-architecture by modifying the output channel dimension of each block, as well as the number of operations or layers of the network. Channel dimension can also be interpreted as the number of filters of convolutional operators within the block, which decides the type and amount of information to extract from previous feature maps. 
It is important to find the appropriate channel depths especially within an object detection framework since feature maps of multiple levels are utilized for localizing and classifying various scales of objects. In our work, we propose to search the optimal channel dimensions of the source backbone network, while considering the trade-off between the computational cost and accuracy of the given detection framework. 
We summarize our main contributions as follows:

\begin{itemize}
\item 
We propose to adapt the source backbone within an object detection framework by searching for the optimal operations and output channel dimensions, while enabling the mapping of ImageNet pre-trained parameters in the searching and fine-tuning stages.
\item
We propose an efficient method for channel search which uses a shared block for all channel dimension candidates with a non-overlapping channel masking technique, enabling fast search and creating a decoupling effect among different channel dimension candidates.
\item 
Our method outperforms other state-of-the-art handcrafted and searched backbones in object detection accuracy. 
\end{itemize}

\begin{figure*}[t]
\centering
\includegraphics[width=1.0\linewidth]{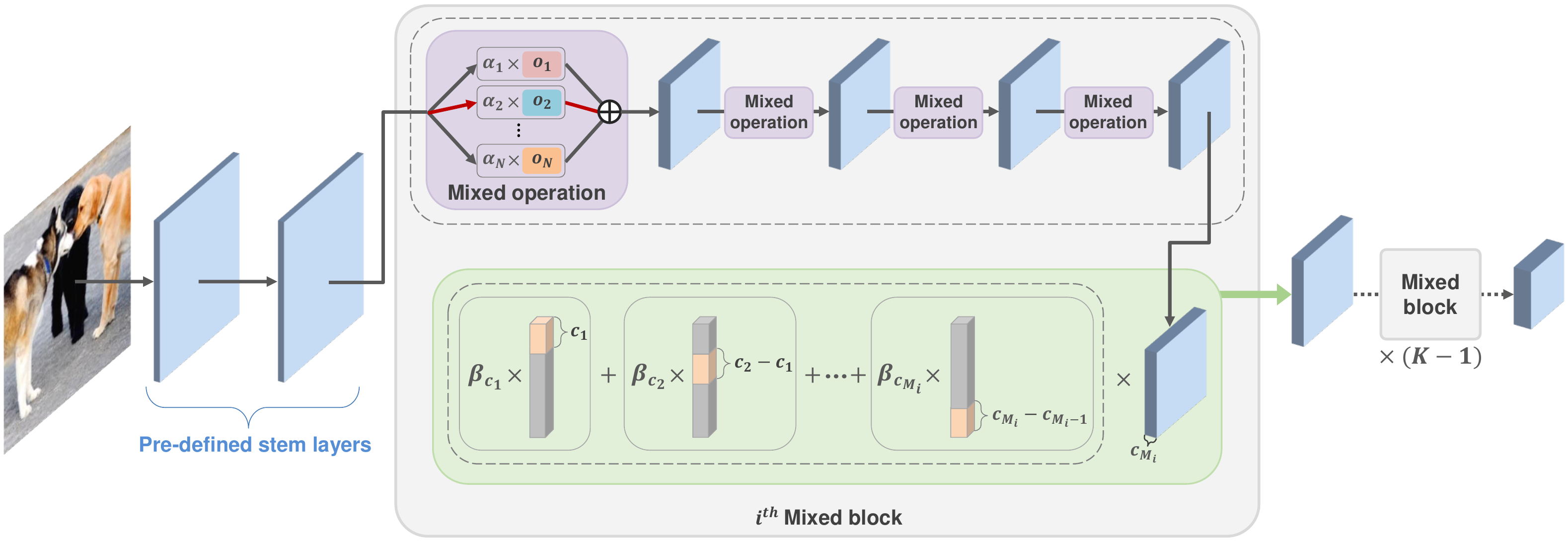}
\caption{Illustration of the overall supernet. The supernet expresses the whole search space with $K$ searchable blocks, where each block has a total of $M$ output channel dimension candidates and $n$ searchable operations with $N$ operation candidates. A mixed operation is defined as a weighted sum of all operation candidates and a mixed block is defined as an element-wise multiplication between a weighted sum of non-overlapping channel mask vectors and one shared feature map with the maximum channel dimension candidate $c_{M_i}$. After the supernet is sufficiently trained, the operation with the highest $\alpha$ and the channel dimension with the highest $\beta$ is chosen to form the final searched architecture.}
\label{fig:1}

\end{figure*}

\section{RELATED WORK}

\subsection{Neural Architecture Search Methods}
Neural Architecture Search (NAS) aims at automatically designing neural network architectures. The main issue of NAS is to decide what to search for and how to search for those components. Early works have used reinforcement learning (RL) \cite{chen2020mnasfpn, xu2019auto, ghiasi2019fpn, xiong2021mobiledets, zhong2018practical} and evolutionary algorithm (EA) \cite{chen2019detnas, guo2020powering, real2017large, elsken2019efficient} as the search strategy. However, these methods often require huge computational costs in the searching stage, usually thousands of GPU days. Later on, ENAS \cite{pham2018efficient} proposed a more efficient approach of RL by allowing child models to share parameters, eliminating the need for training from scratch every time. In recent years, many efficient one-shot approaches \cite{liu2018darts, guo2020powering} were proposed. By introducing the idea of supernet, which is a representation of all possible architectures in the search space, they dramatically reduced the overall search cost. In particular, differentiable NAS \cite{liu2018darts, peng2019nats, fang2020densely, fang2020fna} enabled fast search by training with stochastic gradient descent. Unlike previous approaches that used evolutionary or reinforcement learning over a discrete and non-differentiable search space, differentiable NAS implemented a continuous relaxation of the search space. We also use a differentiable search strategy to efficiently carry out the proposed architecture adaptation.

\subsection{Neural Architecture Search for Object Detection}
Object detection is one of the fundamental tasks in computer vision and it aims to both localize and  classify each  object instance  given  an  image. Two-stage detectors, such as R-CNN \cite{girshick2014rich}, Fast-RCNN \cite{girshick2015fast}, and Faster R-CNN \cite{ren2015faster}, usually have a separate region proposal network (RPN) that selects anchor boxes likely to include object instances before classifying them, whereas one-stage detectors, such as SSD \cite{liu2016ssd}, YOLO \cite{redmon2016you}, and RetinaNet \cite{lin2017focal}, perform localization and classification at once. In general, object detectors consist of three components: a backbone that extracts features from the input image, a neck attached to the backbone that fuses the features, and a head that localizes and classifies object instances from the extracted features. 

Previous NAS literature usually selected one component to search for out of the three parts. DetNAS \cite{chen2019detnas} was the first work to search for a backbone specifically for object detection. DetNAS first pre-trained the supernet with ImageNet \cite{deng2009imagenet} and fine-tuned it on COCO \cite{lin2014microsoft}. Then, the architecture search on the trained supernet was conducted using EA. MobileDets \cite{xiong2021mobiledets} suggested using RL to find the expansion or compression factor of channels. Neck search algorithms mostly aim to find a novel architecture for the feature pyramid network (FPN). For instance, NAS-FPN \cite{ghiasi2019fpn} attempted to find a FPN architecture in a novel scalable search space covering all cross-scale connections. Head search algorithms aim to find optimal sub-networks for both classification and localization. NAS-FCOS \cite{wang2020fcos}, in addition to searching for the FPN, aimed to search the prediction head of an anchor-free object detector named FCOS. Moreover, Hit-Detector \cite{guo2020hit} proposed a hierarchical trinity search to simultaneously search all three components including the backbone, neck, and head. They argued that it is suboptimal to search for each part separately, as they essentially communicate with each other throughout the overall framework. 

It is important to consider the fact that backbones for object detection require ImageNet pre-training for faster convergence and higher accuracy in most cases. The cost of pre-training is non-negligible since ImageNet is a huge dataset. In order to utilize the pre-trained parameters, architecture adaptation or transformation methods for object detection were proposed. NATS \cite{peng2019nats} and Liu \emph{et al}. \cite{liu2021inception} proposed to adjust the dilation rates of convolution operators at the channel level. 
FNA++ \cite{fang2020fna} proposed to adjust kernel sizes and expansion ratios of convolutional blocks. While FNA++ \cite{fang2020fna} focuses on modifying the micro-architecture (building blocks, operations, etc.), we propose to also modify the macro-architecture (number of layers, channel dimensions, etc.).

\section{METHOD}

\subsection{Neural Architecture Search}
\subsubsection{Search Space}
We expand the source network to form a supernet, which represents the overall search space as shown in Fig. \ref{fig:1}. Since the source network is a hand-crafted architecture originally designed for the classification task, the later layers for classification are disregarded. Moreover, the first two blocks are identical to the source network. These pre-defined stem layers are followed by $K$ searchable blocks. 
In this work, we experiment with MobileNetV2  \cite{sandler2018mobilenetv2} as the source network within the RetinaNet framework \cite{lin2017focal}, but our search method can be generally applied to other one-stage or two-stage detectors, as well as other backbone networks. The overall search space is described in detail in Table \ref{table:1}.

\begin{figure}[ht!]
    \centering
    \includegraphics[width=0.5\linewidth]{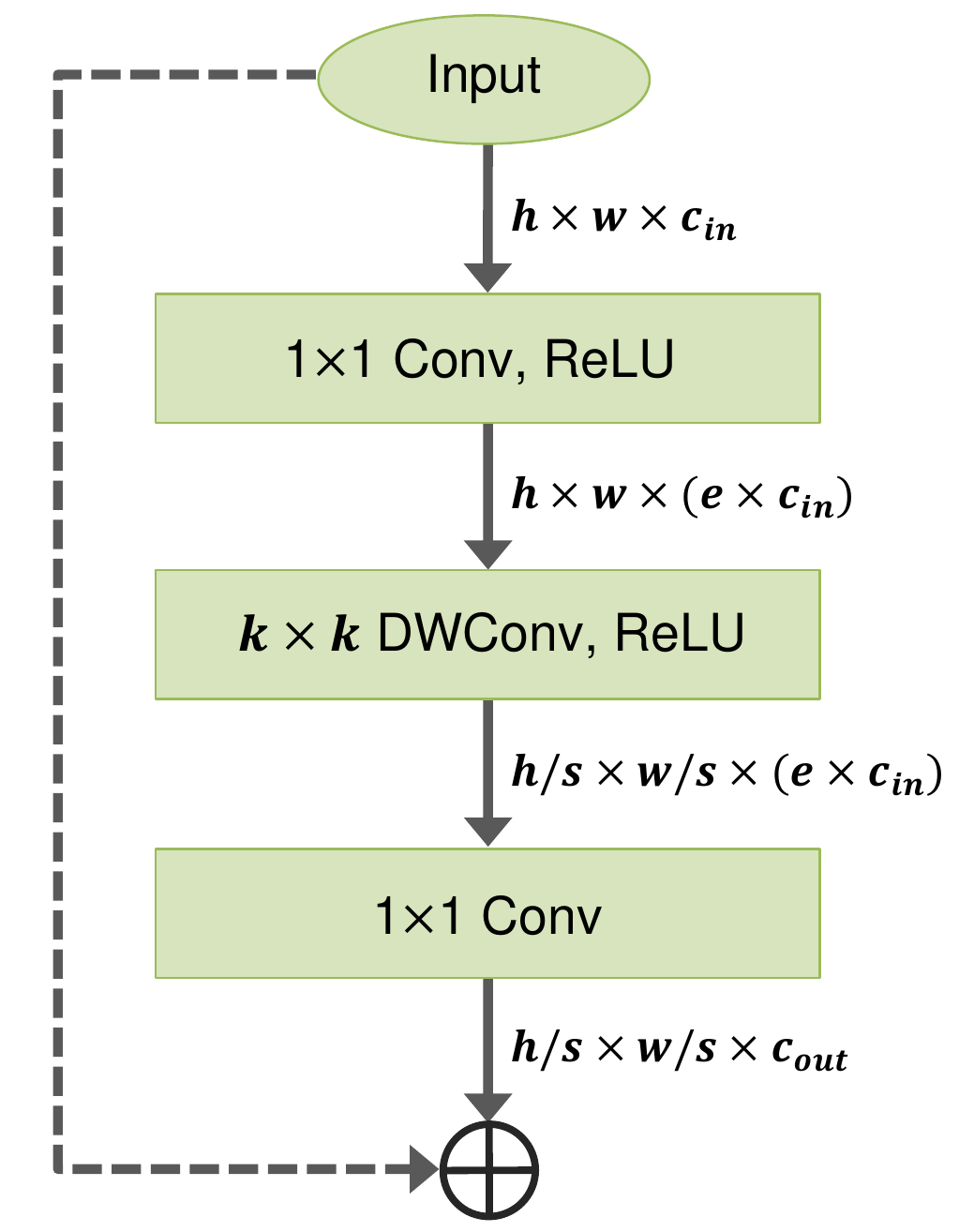}
    \caption{MBConv operation. The search space includes operation candidates with various kernel sizes $k$, expansion factors $e$, and output channel dimensions $c_{out}$.}\label{fig:3}

\end{figure}

\textbf{Operation candidates.} 
We define $N$ operation candidates, which include variations of the inverted residual convolution modules (MBConv) of MobileNetV2 \cite{sandler2018mobilenetv2} and skip connection. MBConv operation includes a sequence of layers, as shown in Fig. \ref{fig:3}. More specifically, the search space includes MBConv operations with kernel sizes of $\{3, 5, 7\}$ and expansion factors of $\{3, 6\}$, consistent for all blocks. Kernel sizes control the distribution of effective receptive fields, which is especially an important factor for detecting various scales of objects, while expansion factors control how much the channels are expanded and reduced within each operation.
The blocks of MobileNetV2 \cite{sandler2018mobilenetv2} include at most 4 operations, so we allow up to 4 operations to be searched in each block for further flexibility. When skip connection is chosen, the corresponding operation is removed, adjusting the total number of layers. In order to avoid the case where all operations are skipped within each block, we exclude the skip connection candidate for the first MBConv operation.

\setlength{\tabcolsep}{5pt} 
\renewcommand{\arraystretch}{1.2} 
\begin{table}[t]
\caption{Search space based on MobileNetV2 \cite{sandler2018mobilenetv2}. "SB" denotes searchable block, $n$ denotes the number of MBConv operations and $s$ indicates the stride of the first operation within each block. $e$ stands for expansion ratio and $k$ stands for kernel size. For output channel dimension $c_{out}$, the tuples of three elements represent (minimum, maximum, step).}
\label{table:1}
    \begin{center}
    \begin{tabular}{c|ccccc}
    \hline
    Block type & $n$ & $s$ & $k$ & $e$ & $c_{out}$ \\
    \hline
    2D Conv & 1 & 2 & 3 & - & 32\\
    MBConv ($k3\_e1)$ & 1 & 1 & 3 & 1 & 16\\
    TBS & 4 & 2 & $\{3,5,7\}$ & $\{3,6\}$ & (16, 28, 2)\\
    TBS & 4 & 2 & $\{3,5,7\}$ & $\{3,6\}$ & (28, 48, 2)\\
    TBS & 4 & 2 & $\{3,5,7\}$ & $\{3,6\}$ & (48, 72, 2)\\
    TBS & 4 & 1 & $\{3,5,7\}$ & $\{3,6\}$ & (72, 128, 4)\\
    TBS & 4 & 2 & $\{3,5,7\}$ & $\{3,6\}$ & (128, 256, 4)\\
    TBS & 1 & 1 & $\{3,5,7\}$ & $\{3,6\}$ & (256, 400, 4)\\
    \hline
    \end{tabular}
    \end{center}

\end{table}

\textbf{Channel dimension candidates.}
We define $M_i$ output channel dimension candidates for each block $B_i$ and the specific set of candidates differ for each block. A block contains multiple MBConv operations with the same output channel dimension. The output channels may either increase, decrease or stay the same compared to the input dimension. 
The output channel dimensions of the first two blocks are fixed, following the source network. The channel dimensions of the source network MobileNetV2 \cite{sandler2018mobilenetv2} for the 6 searchable blocks are $[32, 16, 24, 32, 64, 96, 160, 320]$ and the specific channel candidates are described in Table \ref{table:1}. Empirically, we found that keeping or increasing the channel dimensions for subsequent blocks throughout the network is most effective.

\begin{figure*}[t]
    \centering
    \includegraphics[width=1.0\linewidth]{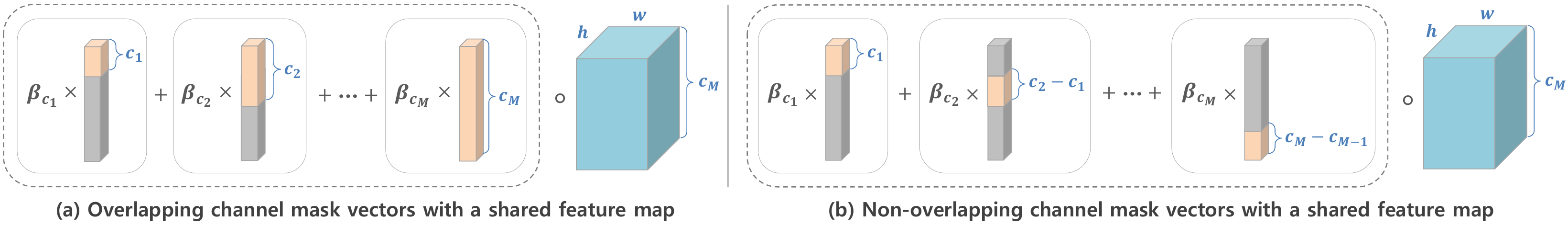}
    \caption{Mixed block with overlapping channel masks and the proposed non-overlapping channel masks. There are $M$ candidate blocks, each with different output channel dimensions up to $c_{M}$ with beta parameters up to $\beta_{c_M}$. $w$ and $h$ denote the width and height of the shared feature map.}\label{fig:2}

\end{figure*}

\subsubsection{Continuous Relaxation of Search Space}
In order to make the search space differentiable, it is necessary to relax the discrete supernet to be continuous. We achieve this using the concept of a mixed operation and a mixed block.

\textbf{Mixed operation.}
Let $\mathbb{O}$ be the set of operation candidates in the search space. In the supernet, we assign an architecture parameter $\alpha_{o}$ to each candidate operation $o \in \mathbb{O}$. Following DARTS \cite{liu2018darts}, in order to make the discrete architecture search space into a continuous one, we relax the discrete choice of a specific operation to a softmax function over all operation candidates. By passing $\alpha$ through softmax, each operation is assigned a probability value. The mixed operation of the $\ell$th operation is defined as a weighted sum of all operation candidates and can be expressed as:
\begin{equation}\label{eq:1}
    \overline{o}_{\ell}(x_{\ell-1}) = \sum_{o \in \mathbb{O}}
    {\frac{\exp{(\alpha_o)}}{\sum_{o^{\prime} \in \mathbb{O}} \exp{(\alpha_{o^{\prime}})}} o(x_{\ell-1})},
\end{equation}
where $x_{\ell-1}$ denotes the input tensor, which is the output feature map of the previous operation $\overline{o}_{\ell-1}$. At the end of search, a discrete architecture can be obtained by replacing each mixed operation $\overline{o}_{\ell}$ with the most likely operation, that is, the operation with the highest $\alpha$ parameter.

\textbf{Mixed block.}
In a similar fashion to how we defined mixed operation, we define a mixed block as a weighted sum of all candidate blocks with different output channel dimensions. In this work, a block is defined as a sequence of mixed operations with the same output channel dimensions. Let $\mathbb{C}$ be the set of output channel dimension candidates of the $i$th mixed block. In the supernet, we assign an architecture parameter $\beta_c$ to each candidate $c \in \mathbb{C}$. The $i$th mixed block can be expressed as:
\begin{equation}\label{eq:2}
    \begin{aligned}
        &\overline{B}_{i}[\mathbb{C}, n, s](x_{i-1}) \\
        &= {B}_{i}[c_{M_i}, n, s](x_{i-1}) \circ \sum_{c \in \mathbb{C}}
        {\frac{\exp{(\beta_{c})}}{\sum_{c^{\prime} \in \mathbb{C}} \exp{(\beta_{c^{\prime}})}}} \mathbb{M}_{c},
    \end{aligned}
\end{equation}
where the inputs of $B_i$ are $n$ that denotes the number of mixed operations and $s$ that refers to the stride of the first operation within the block. The strides of the following operations are set to 1. The specific set of candidates differ for each mixed block, having up to $M_i$ candidates. $\mathbb{M}_c$ denotes the zero mask column vector corresponding to each channel dimension candidate $c$. 

A naive approach would be creating individual blocks for each output channel candidate and adding them through zero-padding, as the output feature maps originally cannot be added due to mismatching channel dimensions. However, this method incurs linearly increasing cost as more channel candidates are added. Therefore, we suggest a more efficient approach of using one shared block that has the largest channel dimension $c_{M_i}$ and channel masking vectors, which only requires one forward pass and one feature map. One possible approach is to simply mask out the extra channels exceeding the current channel candidate, as shown in Fig. \ref{fig:2} (a). However, even when the weight of a higher channel dimension becomes very high, its value cannot easily overcome the accumulated weights of lower dimensions and as a result, the supernet was not properly trained. Therefore, we propose to use non-overlapping channel mask vectors, as shown in Fig. \ref{fig:2} (b). This approach creates a decoupling effect among different channel candidates. When the weighted sum of channel masks is multiplied to the shared block through Hadamard product, the resulting mixed block embeds the various weights of different channel options. In other words, the channel dimensions with higher $\beta$ weights would be clearly emphasized within the mixed block, while those with lower $\beta$ weights would have less effect.

\setlength{\tabcolsep}{10.5pt} 
\renewcommand{\arraystretch}{1.3} 
\begin{table*}[t]
\caption{Comparison of object detection results of RetinaNet \cite{lin2017focal} on COCO test-dev \cite{lin2014microsoft}. \\ Network type denotes whether the backbone was hand-crafted or searched through NAS. \\
For baseline models, we directly cite the search cost based on the GPU used in their original papers. \\
}
\label{table:2}
\begin{center}
\begin{tabular}{l|c|l|cc|c|c}
\hline
Detector & Network Type & Backbone & Params & MAdds & \thead{mAP \\ (\%)} & \thead{Search Cost \\ (GPU-days) }\\
\hline
\multirow{9}{*}{RetinaNet} & \multirow{2}{*}
{Hand-crafted} & MobileNetV2 \cite{sandler2018mobilenetv2} & 11.49M & 133.05B & 32.8 & -\\
&& ResNet-50 \cite{he2016deep} & 37.97M& 202.84B  & 35.5 & -\\
\cline{2-7}
&\multirow{7}{*}
{NAS} & DetNAS \cite{chen2019detnas} & 13.41M & 133.26B & 33.3 & 44.0$^*$\\
&& FNA \cite{fang2020fast} & 11.73M & 133.03B & 33.9 & 6.0$^\dagger$\\
&& FNA++ \cite{fang2020fna} & 11.90M & \textbf{132.86B} & 34.7 & 5.3$^\dagger$\\
&& FBNet-C \cite{wu2019fbnet} & 12.65M & 134.17B & 34.9 & 9.0$^\ddagger$\\
&& Proxyless (GPU) \cite{cai2018proxylessnas} & 14.62M & 135.81B & 35.0 & 8.3$^*${}\\
&& DenseNAS-C \cite{fang2020densely} & 13.24M & 133.91B & 35.1 &  2.7$^*$\\
&& Ours & \textbf{11.63M} & 134.92B & \textbf{35.5} & \textbf{1.7}$^\S$\\
\hline
\end{tabular}
\end{center}

\vspace{-1.5mm}
\footnotesize
\hspace{18mm}
$^*$NVIDIA V100. 
$^\dagger$TITAN Xp. 
$^\ddagger$GPU not specified, 
$^\S$NVIDIA A100.
\vspace{-2.5mm}
\end{table*}

\normalsize

\subsubsection{Optimization}
We optimize the supernet by both training the operation and architecture parameters using two sets of training data of the same
size, $trainA$ and $trainB$. We aim for a bi-level optimization with $\alpha$ and $\beta$ as the upper-level variables and $w$ as the lower-level variable:
\begin{equation}\label{eq:4}
    \begin{aligned}
        &\min_{\alpha, \beta}{\mathcal{L}_{trainB}(w^*(\alpha, \beta), \alpha, \beta)}\\
        &\text{s.t.}\;  w^* (\alpha, \beta) = \argmin_w  \mathcal{L}_{trainA}(w, \alpha, \beta).
    \end{aligned}
\end{equation}
At first, we solely train the operation parameters $w$ for some epochs until the model accuracy is not too low. After the operations are sufficiently trained, we start to alternatively train the operation parameters and architecture parameters, which approximates to a bi-level optimization.

Although searching for a model with the highest detection accuracy is important, it is also crucial to consider the computational cost, which decides the speed and efficiency of the detector framework. We use Multiply-Adds operations (MAdds) as the metric for measuring the computational cost. We define the loss function as a multi-objective optimization by adding a cost-based regularization term. This term adds some cost constraint so that an efficient model can be searched. The overall loss function is defined as:
\begin{equation}\label{eq:5}
    \begin{aligned}
        \mathcal{L}(w, \alpha, \beta) = \mathcal{L}_{model}(w, \alpha, \beta) 
        + \lambda C(\alpha, \beta),
    \end{aligned}
\end{equation}
where $\mathcal{L}_{model}$ denotes the loss from the model including the classification loss and regression loss, while $\lambda$ controls the magnitude of the cost regularization.

In order to estimate the model cost during the search stage, we initially create a lookup table recording the cost of each operation in the search space. In a similar fashion to how we define mixed operation and mixed block, we consider the probabilities embedded in the architecture parameters and compute the MAdds of the whole architecture, expressed as:
\begin{align}
    &C(\alpha, \beta) = \sum_{i=1}^{K} \overline{C}^i(\alpha, \beta), \\
    &\overline{C}^i(\alpha, \beta) =  \sum_{c \in \mathbb{C}^i}
    {\frac{\exp{(\beta_{c})}}{\sum_{c^{\prime} \in \mathbb{C}} 
    \exp{(\beta_{c^{\prime}})}}}C^{i}_c(\alpha), \\
    &C^{i}_c(\alpha) = \sum_{\ell=1}^{n} \overline{C}_c^{i,\ell}(\alpha),\\
    &\overline{C}_c^{i,\ell}(\alpha) = \sum_{o \in \mathbb{O}}{\frac{\exp{(\alpha_{o})}}{\sum_{o^{\prime} \in \mathbb{O}} \exp{(\alpha_{o^{\prime}})}}} C_{c, o}^{i,\ell},
\end{align}
where the total cost $C$ is calculated by summing the cost of $K$ mixed blocks, which is a weighted sum of $\beta$ weights and the cost of each candidate block $C^{i}_c$ with channel dimension candidate $c$ as in Eq. 6. In Eq. 7, the cost of the $i$th mixed block is computed by summing the cost of $n$ mixed operations within the block, which is a weighted sum of $\alpha$ weights and the cost of each operation candidate $o$ as in Eq. 8.

\subsection{Mapping Pre-trained Parameters}
After the supernet is trained, we derive the final architecture by selecting the operation with the highest $\alpha$ parameter and the output channel dimension of each block with the highest $\beta$ parameter.
The pre-trained parameters of the source network are mapped to the supernet and the searched architecture. Our method enables parameter mapping for different number of operations, channel dimensions, and kernel sizes.

\subsubsection{Number of operations}
Our method enables depth search by adjusting the number of operations in each block (stage), which ultimately decides the total number of layers of the searched network. Our search space includes the skip connection operator, which removes the corresponding operation in the searched network. The parameters of extra operations or layers are copied from the last layer in the original block.

\subsubsection{Channel dimension}
Our method enables width search by adjusting the number of output channels of each block and the expansion rate of MBConv operations. When a higher channel dimension is chosen, the parameters of corresponding dimensions are directly mapped and additional channels are assigned 0. When a lower channel dimension is chosen, the exceeding channels are disregarded.

\subsubsection{Kernel size}
Our method enables operation search by adjusting the kernel sizes of MBConv operations. When the searched kernel size is larger than that of the source network, the original parameters are mapped in the central region and the surrounding region is assigned 0. When the searched kernel size is smaller, only the overlapping central region is mapped from the source network and the rest are disregarded. We add small random noises to the mapped parameters to allow backpropagation of those assigned 0.


\setlength{\tabcolsep}{4.5pt} 
\renewcommand{\arraystretch}{2.2} 
\begin{table*}[t]
\caption{Comparison between using individual blocks for each channel dimension candidate and using one shared block for all candidates. The tuples of three elements of channel dimension candidates represent the lowest channel dimension, highest channel dimension, and steps within the range. Object detection results of RetinaNet \cite{lin2017focal} on COCO minival \cite{lin2014microsoft} are reported.}
\label{table:3}
    \begin{center}
    \begin{tabular}{c|c|c|c|cc|c}
    \hline
    Method & \thead{Channel dimension candidates} & \thead{Max. number of channel \\ candidates per block} & Search cost & Params & MAdds & mAP ($\%$) \\ 
    \hline
    Individual block & 
    \makecell{$\{24, 32, 64\}, \{32, 64, 96\}, \{64, 96, 160\},$\\ 
    $\{96, 160, 320\}, \{160, 320, 480\}, \{320, 480, 640\}$}
        & 3 & 9.48 GPU-days & 12.74M & 133.84B & 33.8 \\
    \hline
    Shared block & 
    \makecell{$(16, 28, 2), (28, 48, 2), (48, 72, 2),$ \\$(72, 128, 4), (128, 256, 4), (256, 400, 4)$} 
    & 37 & 1.69 GPU-days & 11.63M & 134.92B & 35.3 \\
    \hline
    \end{tabular}
    \end{center}
\vspace{-2.5mm}
\end{table*}

\setlength{\tabcolsep}{3pt} 
\renewcommand{\arraystretch}{1.3} 
\begin{table}[t]
\caption{Comparison between overlapping and non-overlapping channel mask technique for mixed blocks, as well as between training from scratch and fine-tuning from pre-trained parameters. Object detection results of RetinaNet \cite{lin2017focal} on COCO minival \cite{lin2014microsoft} are reported.}
\label{table:4}
    \begin{center}
    \begin{tabular}{l|l|cccccc}
    \hline
    Channel masking & Training & mAP & $\text{AP}_{50}$ & $\text{AP}_{75}$ & $\text{AP}_S$ & $\text{AP}_M$ & $\text{AP}_L$\\
    \hline
    \multirow{1}{*}{Overlapping} & Pre-trained & 26.5 & 42.4 & 27.8 & 14.5 & 27.0 & 35.9\\
    \hline
    \multirow{2}{*}{Non-overlapping} & Scratch & 33.0 & 52.8 & 35.5 & 18.0 & 36.5 & 43.4 \\
    & \textbf{Pre-trained} & \textbf{35.3} & \textbf{54.2} & \textbf{37.3} & \textbf{19.4} & \textbf{38.4} & \textbf{46.1}\\
    \hline
    \end{tabular}
    \end{center}
\vspace{-3mm}
\end{table}

\section{EXPERIMENTS}

\subsection{Implementation Details}
Our model is implemented based on PyTorch \cite{NEURIPS2019_bdbca288}. For RetinaNet \cite{lin2017focal}, the input image is resized to 800$\times$1333 during inference and to 800$\times$1088 for calculating Multiply-Adds operations (MAdds) following previous works. 
The search process runs for 14 epochs, where only operation parameters are trained for the first 8 epochs, and from epoch 9, architecture parameters are activated and trained alternatively with operation parameters.
The search cost takes approximately 1.7 GPU-days on a single NVIDIA A100 GPU with a batch size of 8. 
Operation parameters are trained with the SGD optimizer using a learning rate of 0.02, momentum of 0.9, and weight decay of 0.0001. Architecture parameters are trained with the Adam optimizer using a learning rate of 0.0003 and weight decay of 0.001. 
Lastly, we fine-tune the searched network for 24 epochs, which takes approximately 3.6 GPU-days on NVIDIA A100 GPU with a batch size of 32. We use the SGD optimizer with a learning rate of 0.05, momentum of 0.9, and weight decay of 0.00005.

\subsection{Quantitative Results}
We report results using mean Average Precision (mAP), Average Precision (AP) with IoU threshold of 0.50 and 0.75, AP of small, medium, and large object scales. In addition, we report the number of parameters and the number of MAdds to compare the size and computation cost of various backbones. Table \ref{table:2} reports detection results of RetinaNet \cite{lin2017focal} using various backbones on the COCO test-dev set \cite{lin2014microsoft}. Our searched backbone was able to achieve the state-of-the-art accuracy on mAP, compared to other hand-crafted and searched models. Meanwhile, our model had the lowest number of parameters, enabling efficient training and inference. In addition, our searched model achieved comparable detection performance to ResNet-50 \cite{he2016deep}, despite having approximately 70\% less parameters and 35\% less MAdds. Compared to MobileNetV2 \cite{sandler2018mobilenetv2}, our searched model improved the mAP by 2.5\%, with only 1.3\% more parameters and 1.5\% more MAdds.

\subsection{Ablation Studies}
\subsubsection{Comparison between using individual blocks and one shared block for mixed blocks}
We compare two methods of defining the mixed block, between using a shared block with channel mask vectors and using individual candidate blocks with different channel dimensions through zero padding. In the shared block approach, all channel dimension candidates ultimately share the operation weights. The results reported in Table \ref{table:3} show that the search cost of the shared block method is significantly lower than using individual candidate blocks. The low computational cost of the shared block method allowed us to add significantly more channel dimension candidates to search from, which we believe led to better detection accuracy. Considering the specific channel dimension candidates, we empirically found that gradually increasing the channel depth throughout the network yielded better results.

\subsubsection{Comparison between overlapping and non-overlapping channel masks for mixed blocks}
Table  \ref{table:4} compares the results of two channel masking methods, the method using overlapping mask vectors proposed in FBNetV2 \cite{wan2020fbnetv2}, shown in Fig. \ref{fig:2} (a), and our proposed method using non-overlapping mask vectors, shown in Fig. \ref{fig:2} (b). 
Our proposed non-overlapping channel masking method achieved higher accuracy overall under the same setting. Empirically, the overlapping channel masking method did not train well, as the gradient values of higher channel dimensions for backpropagation progressively became almost identical. This may be due to the fact that the shallow channel dimensions of overlapping mask vectors are summed up to 1 or nearly 1. To avoid this effect, we suggested to use non-overlapping channel masking vectors to clearly emphasize the effect of optimal channel dimension candidates and to create a decoupling effect between each channel dimension candidate.

\subsubsection{Comparison between training with pre-trained parameters and from scratch}
Existing studies on object detection mostly utilize ImageNet pre-trained parameters of the given backbone. However, searching for a completely new architecture makes the use of existing pre-trained models impossible. This results in high computational cost, especially due to the fact that the large supernet must be pre-trained as well for better search results. Table \ref{table:4} demonstrates that fine-tuning based on pre-trained parameters yield better results than training from scratch.
The results prove the importance of our architecture adaptation approach which allows for the mapping of pre-trained parameters.

\section{CONCLUSION}
In this paper, we proposed a function-preserving architecture adaptation method to search an optimal backbone specifically for object detection, beyond image classification. We achieved this by adjusting the number of layers, operations, and channel dimensions, in both the macro- and micro-level. In particular, searching for the optimal output channel dimensions of each block of the network increased its feature representation capability. One of the biggest advantages of our method is the low searching and training cost, achieved through the non-overlapping channel masking approach and the pre-trained parameter mapping scheme for the supernet and the searched backbone.  
We conducted experiments on the single-stage detection framework with our searched backbone and our model outperformed both manually designed and NAS-based state-of-the-art backbones on the COCO dataset.

\bibliographystyle{IEEEtran}
\bibliography{bare_conf}

\end{document}